\documentclass{article}

\usepackage{graphicx}
\usepackage{hyperref}
\usepackage{authblk}

\begin{document}

\title{Using Traceless Genetic Programming for Solving Multiobjective Optimization Problems}

\author{Mihai Oltean and Crina Gro\c san \\
Department of Computer Science\\
Faculty of Mathematics and Computer Science\\
Babe\c s-Bolyai University, Kog\u alniceanu 1\\
Cluj-Napoca, 3400, Romania.\\
mihai.oltean@gmail.com
}

\maketitle

\begin{abstract}

Traceless Genetic Programming (TGP) is a Genetic Programming (GP) variant that is used in the cases where the focus is rather the output of the program than the program itself. The main difference between TGP and other GP techniques is that TGP does not explicitly store the evolved computer programs. Two genetic operators are used in conjunction with TGP: crossover and insertion. In this paper we shall focus on how to apply TGP for solving multi-objective optimization problems which are quite unusual for GP. Each TGP individual stores the output of a computer program (tree) representing a point in the search space. Numerical experiments show that TGP is able to solve very fast and very well the considered test problems.

\end{abstract}

\section{Introduction}

Koza \cite{koza1,koza2} suggested that Genetic Programming (GP) may be used for solving equations. In that approach \cite{koza1} each GP tree represented a potential solution of the problem (equation). The internal nodes of the tree contained mathematical operators (+,--,*,\slash) and leaves contained real constants (randomly generated) \cite{koza1}. Even if this idea is very interesting and appealing, little work has been dedicated to this issue.

In this paper, we explore the possibility of encoding solutions of multiobjective problems as trees rather than as simple real values. The value of the (expression encoded into a) tree will represent the solution of the problem. Since we need the value of the tree rather than the entire tree, we will use a very fast and efficient GP variant called Traceless Genetic Programming.

Traceless Genetic Programming (TGP) \footnote {The source code for TGP is available at \url{https://github.com/mihaioltean/traceless-genetic-programming}} \cite{oltean_tgp1,oltean_tgp2} is a GP \cite{koza1} variant as it evolves a population of computer programs. The main difference between TGP and GP is that TGP does not explicitly store the evolved computer programs. TGP is useful when the trace (the way in which the results are obtained) between input and output is not important. For instance, TGP is useful for solving problems where a numerical output is needed (i.e. solving equations, function optimization). In this way the space used by traditional techniques for storing the entire computer programs (or mathematical expressions in the simple case of symbolic regression) is saved.

TGP-based algorithm is applied for solving five difficult multiobjective test problems: ZDT1 - ZDT4, ZDT6 \cite{deb1,zitzler1,zitzler2} whose definition domain consists of real-valued variables. We employ two variants of TGP. The first variant does not include an archive for storing the nondominated solutions already found. This very fast variant is able to detect in just one second a very good approximation of the Pareto front, but the diversity of the solutions is sometimes poor. A second variant maintains an archive for a better diversity of solutions. In this variant, the distribution along the Pareto front has been significantly improved.

The TGP-based algorithm is compared to Strength Pareto Evolutionary Approach (SPEA) \cite{zitzler1,zitzler2} and Pareto Archive Evolutionary Algorithm (PAES) \cite{knowles1,knowles2}. Results show that TGP is significantly better than the other considered algorithms for most of the test functions. 

The paper is organized as follows: In section \ref{tgp} the Traceless Genetic Programming technique is described in the context of GP problems. A possible field of applications for TGP is described in section \ref{where_tgp}. Section \ref{tgp_moea} describes TGP in the context of multiobjective evolutionary algorithms. The test problems are briefly presented in section \ref{test_problems}. In section \ref{exp} several numerical experiments for solving the multiobjective problems are performed. An improved TGP algorithm with archive is introduced in section \ref{tgp_archive}. Running time of the TGP algorithm is analyzed in section \ref{running_time}. TGP is compared to SPEA and PAES in section \ref{comparison}. Conclusions and further work directions are indicated in section \ref{conclusions}.

\section{Traceless Genetic Programming}\label{tgp}

In this section the TGP technique \cite{oltean_tgp1} is described in the context of Genetic Programming techniques and problems. Later, in section \ref{tgp_moea}, TGP is presented in the context of numerical problems such as solving equations, function optimization and multiobjective function optimization.

\subsection{Prerequisite}

The quality of a GP individual is usually computed by using a set of fitness cases \cite{koza1,koza2}. For instance, the aim of symbolic regression is to find a mathematical expression that satisfies a set of $m$ fitness cases.

We consider a problem with $n$ inputs: $x_{1}$, $x_{2}$, \ldots $x_{n}$ and one output $f$. The inputs are also called terminals \cite{koza1}. The function symbols that we use for constructing a mathematical expression are $F=\{+,-,*,/, sin\}$.

Each fitness case is given as a ($n+1$) dimensional array of real values:\\

\[
v_1^k ,v_2^k ,v_3^k ,...,v_n^k ,f_k 
\]

\noindent
where $v_j^k $ is the value of the $j^{th}$ attribute (which is $x_{j})$ in 
the $k^{th}$ fitness case and $f_{k}$ is the output for the $k^{th}$ fitness 
case.

More fitness cases (denoted by $m$) are usually given and the task is to find the expression that best satisfies all these fitness cases. This is usually done by minimizing the difference between what we have obtained and what we should obtain:

\[
Fitness = \left\| {\left( {{\begin{array}{*{20}c}
 {o_j^1} \hfill \\
 {o_j^2} \hfill \\
 {o_j^3} \hfill \\
 {...} \hfill \\
 {o_j^m} \hfill \\
\end{array} }} \right) - \left( {{\begin{array}{*{20}c}
 {f_1} \hfill \\
 {f_2} \hfill \\
 {f_3} \hfill \\
 {...} \hfill \\
 {f_m} \hfill \\
\end{array} }} \right)} \right\|.
\]

By linearizing the formula above we obtain:

\[
Q=\sum\limits_{k = 1}^m {\left| {f_k - o_k } \right|} ,
\]

\noindent
where $f_{k}$ is the target value for the $k^{th}$ fitness case and $o_{k}$ is 
the actual (obtained) value for the $k^{th}$ fitness case.

\subsection{Individual representation}

Each TGP individual represents a mathematical expression evolved so far, but 
the TGP individual does not explicitly store this expression. Each TGP 
individual stores only the value already obtained for each fitness case. Thus 
a TGP individual is:\\

\[
\left( {{\begin{array}{*{20}c}
 {o_k^1} \hfill \\
 {o_k^2} \hfill \\
 {o_k^3} \hfill \\
 {...} \hfill \\
 {o_k^m} \hfill \\
\end{array} }} \right)
\]

\noindent
where $o_{k}$ is the current value for the $k^{th}$ fitness case. Each 
position in this array (a value $o_{k})$ is a gene. As said before, 
behind these values lies a mathematical expression whose evaluation has 
generated these values. However, we do not store this expression. We only store 
the values $o_{k}$.\\

\textbf{Remark}

\begin{itemize}

\item[{\it (i)}]{The structure of a TGP individual can be easily improved so that it could store the evolved computer program (mathematical expression). Storing the evolved expression can provide an easier way to analyze the results of the numerical experiments. However, in this paper, we shall not store the trees associated with the TGP individuals.}

\item[{\it (ii)}]{TGP cannot be viewed as a GP technique with linear representation \cite{oltean_complex} mainly because TGP does not explicitely store the entire computer program (the tree).}

\end{itemize}

\subsection{Initial population}

The initial population contains individuals whose values have been generated 
by simple expressions (made up of a single terminal). For instance, if an 
individual in the initial population represents the expression:\\

$E=x_{j}$,\\

\noindent
then the corresponding TGP individual is represented as:\\

\[
\left( {{\begin{array}{*{20}c}
 {v_j^1 } \hfill \\
 {v_j^2 } \hfill \\
 {v_j^3 } \hfill \\
 {...} \hfill \\
 {v_j^m } \hfill \\
\end{array} }} \right)
\]

\noindent
where $v_j^k $ has been previously explained.

The quality of this individual is computed by using the equation previously described:

\[
Q = \sum\limits_{i = 1}^m {\left| {v_1^k - f_k } \right|} .
\]

\subsection{Genetic Operators}

The genetic operators which are used in conjunction with TGP are 
described in this section. TGP uses two genetic operators: crossover and insertion. These operators are specially designed for the TGP technique.

\subsubsection{Crossover}\label{Crossover}

The crossover is the only variation operator that creates new individuals. Several individuals (the parents) and a function symbol are selected for 
crossover. The offspring is obtained by applying the selected 
operator to each of the genes of the parents.

\begin{figure*}[htbp]
\centerline{\includegraphics[width=\textwidth]{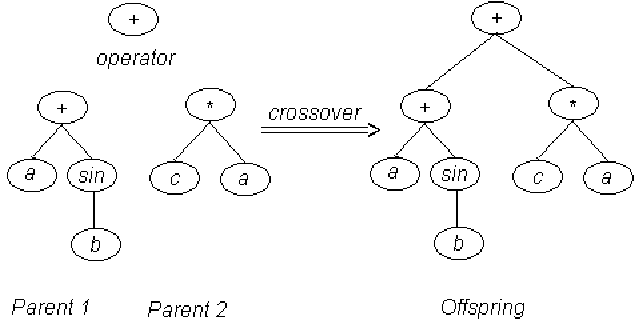}}
\caption{An example of TGP crossover.}
\label{tgp_cross}
\end{figure*}

Speaking in terms of expressions and trees, an example of TGP crossover is depicted in Figure \ref{tgp_cross}.

From Figure \ref{tgp_cross} we can see that the parents are subtrees of the offspring.

The number of parents selected for crossover depends on the number of 
arguments required by the selected function symbol. Two parents have to be selected for crossover if the function symbol is a binary operator. A single parent has to be selected if the function symbol is a unary operator.\\

\textbf{Example 1}\\

Let us suppose that the operator + is selected. In this case two parents are selected and the offspring $O$ is obtained as follows:\\

\[
\left( {{\begin{array}{*{20}c}
 {p_j^1 } \hfill \\
 {p_j^2 } \hfill \\
 {p_j^3 } \hfill \\
 {...} \hfill \\
 {p_j^m } \hfill \\
\end{array} }} \right)\,\,\,operator + \,\,\left( {{\begin{array}{*{20}c}
 {q_j^1 } \hfill \\
 {q_j^2 } \hfill \\
 {q_j^3 } \hfill \\
 {...} \hfill \\
 {q_j^m } \hfill \\
\end{array} }} \right)\buildrel {Crossover} \over \longrightarrow \left( 
{{\begin{array}{*{20}c}
 {p_j^1 + q_j^1 } \hfill \\
 {p_j^2 + q_j^2 } \hfill \\
 {p_j^3 + q_j^3 } \hfill \\
 {...} \hfill \\
 {p_j^m + q_j^m } \hfill \\
\end{array} }} \right)
\]

\noindent

\textbf{Example 2}\\

Let us suppose that the operator \textit{sin} is selected. In this case one parent is selected and the offspring $O$ is obtained as follows:\\

\[
\left( {{\begin{array}{*{20}c}
 {p_j^1 } \hfill \\
 {p_j^2 } \hfill \\
 {p_j^3 } \hfill \\
 {...} \hfill \\
 {p_j^m } \hfill \\
\end{array} }} \right)\,\,\,operator\,\,\sin \buildrel {Crossover} \over 
\longrightarrow \left( {{\begin{array}{*{20}c}
 {\sin (p_j^1 )} \hfill \\
 {\sin (p_j^2 )} \hfill \\
 {\sin (p_j^3 )} \hfill \\
 {...} \hfill \\
 {\sin (p_j^m )} \hfill \\
\end{array} }} \right)
\]

\textbf{Remark} Standard GP crossover \cite{koza1} cannot be simulated within TGP mainly because TGP does not store the entire tree and thus no subtree can be extracted.

\subsubsection{Insertion}\label{insertion}

This operator inserts a simple expression (made up of a single terminal) in the population. This operator is useful when the population contains individuals representing very complex expressions that cannot improve the search. By inserting simple expressions we give a chance to the evolutionary process to choose another direction for evolution. Insertion is very similar to a restart operation.

\subsection{Constants within TGP system}\label{constants}

Some numerical constants might appear within the solution when solving symbolic regression problems. Constants are handled as any other variable. There are many ways to insert constants within TGP. One can use the same value for all the fitness cases, or one can generate different random values for each fitness case. An example of a constant chromosome is given below (values have been randomly generated):

\[
\left( {{\begin{array}{*{20}c}
 {0.7 } \hfill \\
 {2.1 } \hfill \\
 {0.001 } \hfill \\
 {...} \hfill \\
 {3.4 } \hfill \\
\end{array} }} \right)
\]

The use of constants is important in solving real-world classification and symbolic regression problems \cite{prechelt1,uci} where the evolved expression is more than a simple combination of inputs \cite{brameier1}.

The use of constants becomes even more important in the case of function optimization (see the sections below) since, in this case, we don't have any input variables (as those supplied in the case of symbolic regression or classification). In this case all the solutions are made up of (combinations of) random numbers.

\subsection{TGP Algorithm}\label{algorithm}

TGP uses a special generational algorithm by using two populations, which is given below:

The TGP algorithm starts by creating a random population of individuals. The evolutionary process is run for a fixed number of generations. At the beginning of a generation the best individual is automatically copied into the next population. Then the following steps are repeated until the new population is filled: we may either choose to perform crossover between two existing individuals or to insert a randomly generated individual. Thus, with a probability $p_{insert}$ we generate an offspring made up of a single terminal (see the Insertion operator). With a probability 1-$p_{insert}$ select two parents using a standard selection procedure. The parents are recombined in order to obtain an offspring. The offspring enters the population of the next generation.

\begin{figure*}[htbp]
\centerline{\includegraphics[width=\textwidth]{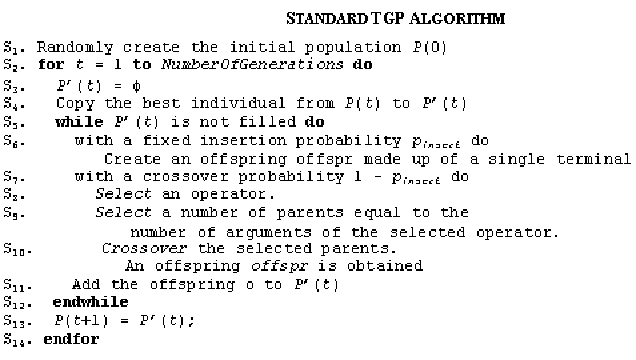}}
\caption{Traceless Genetic Programming Algorithm.}
\label{tgp_alg}
\end{figure*}

The standard TGP algorithm is depicted in Figure \ref{tgp_alg}.

\subsection{Complexity of the TGP Decoding Process} \label{complexity}

A very important aspect of the GP techniques is the time complexity of the procedure used for computing the fitness of the newly created individuals.

The complexity of that procedure for the standard GP is: \\

$O(m*g)$,\\

where $m$ is the number of fitness cases and $g$ is average number of nodes in the GP tree.

By contrast, the TGP complexity is only \\

$O(m)$ \\

because the quality of a TGP individual can be computed by traversing it only once. The length of a TGP individual is $m$.

Due to this reason we may allow TGP programs to run $g$ times more generations in order to obtain the same complexity as the standard GP.

\subsection{Limitations of the TGP}\label{tgp_limitations}

Despite its great speed, TGP has some drawbacks which limit its area of applicability.

In a standard GP system, the evolved tree could be used for analyzing new test data. However, in the TGP system the things are a little bit different. TGP purpose is to avoid storing of the evolved trees in order to achieve speed. This means that the evolved solutions cannot be applied for new and unseed test data. An immediate consequence is that both the training and the test must be known at the moment of training.

\subsection{When TGP is useful?}\label{where_tgp}

As stated in section \ref{tgp_limitations}, TGP cannot be applied for solving symbolic regression or classification problems as we know them. Both the training and the test must be known at the moment of training. And, this is a hard constrain which reduce the space of applicability. After training, TGP cannot be used for analyzing new and unseen data.

But, TGP can be used in other, more subtle, applications:

For instance when we apply a standard Genetic Programming technique we need to know what is the optimal set of functions and terminals. If we run GP with an improper set of functions or terminals we will need a lot of generations and a big population. 

We can use TGP for tuning the set of functions/terminals. Before running the standard GP we could perform many runs with TGP in order to find a good set of functions/terminals. The TGP will not give us a tree or a mathematical expression, but it will give us, very quickly, an indication on whether the selected set of functions and terminals can lead to a good solution of that problem if we apply standard GP.

Note that TGP is very fast. It solved \cite{oltean_tgp1} the even-5-parity problem in 3.2 seconds with no asm or other tricks. This time is infinitely smaller than the time required by a standard GP system to solve this problem. Thus, TGP can be run multiple times in order to see if a set of functions/terminals is good enough for obtaining a good solution with standard GP.

TGP can act as a kind of very fast preprocessing. TGP will not output a tree, but it will tell you very fast if the selected parameters for your algorithm are good or you have to change them before running the entire standard GP system.
	
We can significantly improve the speed of TGP if we use:

\begin{itemize}

\item{the Submachine Code GP \cite{poli2}. The authors offered that SubMachine Code improved the speed by 1.8 orders of magnitude. We expect to improve the speed of TGP by a similar order of magnitude.}

\item{implementation in machine language as in the case of Discipulus \cite{discipulus}. The authors offered that Discipulus is sixty to two hundreds faster than standard Genetic Programming.}

\end{itemize}

Other potentially interesting applications for TGP are currently under consideration: solving single and multi-objective optimization problems and solving equations.

\section{TGP for multiobjective problems}\label{tgp_moea}

The following modifications have been applied to the Traceless Genetic Programming algorithm in order to deal with specific multiobjective problems:

\begin{itemize}
\item[{\it(i)}]{In the case of the considered multiobjective problems \cite{deb1} we do not have a training set as in the case of GP. The TGP chromosome will still be an array of values, but each value will represent a variable of the search space. Thus, the chromosome length will be equal to the number of variables of the considered multiobjective test problem. In our test problems \cite{deb1} each variable will usually hold a value in a real interval (e.g. [0,1], or [-5,5]).} This modification is more a formal than a structural one.

\item[{\it (ii)}]{The initial population will consist of vectors of constants (see section \ref{constants}). Each value in these arrays will be randomly generated over the definition domain of the problem being solved. The subsequent populations will be combinations of these constants or will be replaced, by insertion (see section \ref{insertion}), with some other vectors of constants.}

\item[{\it (iii)}]{The fitness of a TGP chromosome for multiobjective problems is computed by applying it to the considered test problems. For each objective we will have a fitness function. This is different from the standard TGP where we had a single fitness which was computed as a sum over all fitness cases.}

%\item[{\it(ii)}]{There is formal modification of the meaning of chromosomes for the considered MO problems. We said before %that a TGP chromosome will store only the output (the value) of an expression

\item[{\it (iv)}]{All the nondominated solutions from the current generation will be copied into the next generation. This is again different from the standard TGP where only the best solution in the current population was copied into the next generation.}

\end{itemize}

\subsection{TGP algorithm for multiobjective optimization problems}\label{tgp_moea_alg}

The algorithm used for solving multiobjective optimization problems is very similar to those used in the context of GP problems (see section \ref{algorithm}). The main difference is that all the nondominated solutions from the current population are automatically copied into the next generation. The algorithm may be described as follows:

The initial population is randomly generated and the entire evolutionary process is run for a fixed number of generations. At each generation the nondominated solutions from the current population are automatically copied into the next population. Then, the following steps are repeated until the new population is filled: we may either choose to perform crossover between two existing individuals or to insert a randomly generated individual. Thus, with a probability $p_{insert}$ generate an offspring made up of a single terminal (see the Insertion operator). With a probability 1-$p_{insert}$ select parents using a standard selection procedure. The parents are recombined (see section \ref{Crossover}) in order to obtain an offspring. The offspring enters the new population. At the end of the generation, the new population is copied into the old one.

\section{Test problems}\label{test_problems}

The five test functions \cite{deb1,zitzler2} used in this paper for numerical experiments are described in 
this section.

Each test function is built by using three 
functions $f_{1}$, $g$, $h$. Biobjective function $T$ considered here is: \\

$T(x)$ = ($f_{1}(x)$, $f_{2}(x))$. \\

The optimization problem is:

\[
\left\{ {{\begin{array}{*{20}c}
 {Minimize\,T(x),\,\,where\,f_2 (x) = g(x_2 ,\ldots ,x_m )h(f_1 (x_1 
),\,g(x_2 ,\ldots x_m )),} \\
 {x = (x_1 ,\ldots ,x_m )} \\
\end{array} }} \right.
\]

\subsection{Test function ZDT1}

Test function ZDT$_{1}$ is defined by using the following 
functions: \\

$\begin{array}{l}
 f_1 (x_1 ) = x_1 , \\ 
 g(x_2 ,\ldots .x_m ) = 1 + 9 \cdot \sum\nolimits_{i = 2}^m {{x_i } 
\mathord{\left/ {\vphantom {{x_i } {(m - 1)}}} \right. 
\kern-\nulldelimiterspace} {(m - 1)}} , \\ 
 h(f_1 ,g) = 1 - \sqrt {{f_1 } \mathord{\left/ {\vphantom {{f_1 } g}} 
\right. \kern-\nulldelimiterspace} g} , \\ 
 \end{array}$

where $m$ = 30 and $x_{i}$\textit{ $ \in $ }[0,1] $i$ = 1,2,\ldots ,$m$.\\

Pareto optimal front for the problem ZDT$_{1}$ is convex and is 
characterized by the equation $g(x)$ = 1.

\subsection{Test function ZDT2}

Test function ZDT$_{2}$ is defined by considering the 
following functions:\\

$\begin{array}{l}
 f_1 (x_1 ) = x_1 \\ 
 g(x_2 ,\ldots .x_m ) = 1 + 9 \cdot \sum\nolimits_{i = 2}^m {{x_i } 
\mathord{\left/ {\vphantom {{x_i } {(m - 1)}}} \right. 
\kern-\nulldelimiterspace} {(m - 1)}} \\ 
 h(f_1 ,g) = 1 - \left( {{f_1 } \mathord{\left/ {\vphantom {{f_1 } g}} 
\right. \kern-\nulldelimiterspace} g} \right)^2 \\ 
 \end{array}$

where $m$ = 30 and $x_{i} \quad  \in $ [0,1], $i$ = 1,2,\ldots ,$m$.\\

Pareto optimal front is characterized by the equation $g(x)$=1.

ZDT$_{2}$ is the nonconvex counterpart to ZDT$_{1}$.

\subsection{Test function ZDT3}

Pareto optimal set corresponding to the test function 
ZDT$_{3}$ presents a discrete feature. Pareto optimal front consists of 
several noncontiguous convex parts. The involved functions are:\\

$\begin{array}{l}
 f_1 (x_1 ) = x_1 \\ 
 g(x_2 ,\ldots .x_m ) = 1 + 9 \cdot \sum\nolimits_{i = 2}^m {{x_i } 
\mathord{\left/ {\vphantom {{x_i } {(m - 1)}}} \right. 
\kern-\nulldelimiterspace} {(m - 1)}} \\ 
 h(f_1 ,g) = 1 - \sqrt {{f_1 } \mathord{\left/ {\vphantom {{f_1 } g}} 
\right. \kern-\nulldelimiterspace} g} - \left( {{f_1 } \mathord{\left/ 
{\vphantom {{f_1 } g}} \right. \kern-\nulldelimiterspace} g} \right)\sin 
\left( {10\pi f_1 } \right) \\ 
 \end{array}$

where $m$ = 30 and $x_{i} \quad  \in $ [0,1], $i$ = 1,2,\ldots $m$.\\

Pareto optimal front is characterized by the equation $g(x)$ = 1.

\subsection{Test function ZDT4}

The introduction of the function \textit{sin} in the expression of function $h$ causes 
discontinuity in the Pareto optimal front. However, there is no 
discontinuity in the parameter space.

The test function ZDT$_{4}$ contains 21$^{9}$ local Pareto 
optimal fronts and, therefore, it tests the EA ability to deal with 
multimodality. The involved functions are defined by:\\

$\begin{array}{l}
 f_1 (x_1 ) = x_1 \\ 
 g(x_2 ,\ldots .x_m ) = 1 + 10(m - 1) + \sum\nolimits_{i = 2}^m {(x_i^2 - 
10\cos (4\pi x_i ))} \\ 
 h(f_1 ,g) = 1 - \sqrt {{f_1 } \mathord{\left/ {\vphantom {{f_1 } g}} 
\right. \kern-\nulldelimiterspace} g} \\ 
 \end{array}$

where $m$ = 10, $x_{1} \quad  \in $ [0,1] and $x_{2}$\textit{,\ldots ,x}$_{m} \quad  \in $ [-5,5].\\

Global Pareto optimal front is characterized by the equation $g(x)$ = 1.

The best local Pareto optimal front is described by the equation $g(x)$ = 1.25.

Note that not all local Pareto optimal sets are distinguishable in the 
objective space.

\subsection{Test function ZDT6}

The test function ZDT$_{6}$ includes two difficulties caused 
by the nonuniformity of the search space. First of all, the Pareto optimal 
solutions are nonuniformly distributed along the global Pareto optimal front 
(the front is biased for solutions for which $f_{1}(x)$ is a neat one). Secondly, 
the density of the solutions is lowest near the Pareto optimal front and 
highest away from the front. 

This test function is defined by using:\\

$\begin{array}{l}
 f_1 (x_1 ) = 1 - \exp ( - 4x_1 )\sin ^6(6\pi x_1 ) \\ 
 g(x_2 ,\ldots .x_m ) = 1 + 9 \cdot \left( {\sum\nolimits_{i = 2}^m {{x_i } 
\mathord{\left/ {\vphantom {{x_i } {(m - 1)}}} \right. 
\kern-\nulldelimiterspace} {(m - 1)}} } \right)^{0.25} \\ 
 h(f_1 ,g) = 1 - \left( {{f_1 } \mathord{\left/ {\vphantom {{f_1 } g}} 
\right. \kern-\nulldelimiterspace} g} \right)^2 \\ 
 \end{array}$

where $m$ = 10, $x_{i} \quad  \in $ [0,1], $i$ = 1,2,\ldots $m$.\\

The Pareto optimal front is characterized by the equation $g(x)$ = 1, and is nonconvex.

\section{Metrics of performance}\label{metrics}

Both metrics measure the convergence to the Pareto front and can be applied 
only if the Pareto front is known. 

\subsection{Convergence metric}

Assume that the Pareto front is known. Let us denote by $PA$ a set of Pareto 
optimal solutions. For each individual $i$ from the final population \textit{FP} distance 
(Euclidian distance or other suitable distance) $d_{ij}$ to the all points 
$j$ of $P$ is computed.

The minimum distance:

\begin{center}
\textit{mindist}$_{i}=\mathop {\min }\limits_{j \in P} \,d_{ij} $
\end{center}

\noindent
is kept for each individual. 

The average of these distances 

\[
CM = \frac{\sum\limits_{i \in FP} {\mbox{mindist}_{i} } }{\left| {FP} \right|}
\]

\noindent
represents the measure of convergence (the distance) to the Pareto front.\\

\textbf{\it{Remark}}

Lower values of the convergence metric represent good convergence.

\subsection{Diversity metric}

For each individual from the final population \textit{FP} we consider the point from the 
set of Pareto optimal points $P$ situated at minimal distance. 

We called each such point from $PA$ a \textit{marked} point. The total number of different 
marked points from $PA$ over the size of $PA$ represents the diversity metric (\textit{DM}).\\

\textbf{\it{Remark}}

\begin{itemize}
\item[{\it (i)}]{Higher values for \textit{DM} indicate a better diversity.}

\item[{\it (ii)}]{We have considered 200 equidistant points on the Pareto front in order to compute the diversity metric. The value of diversity metric has been normalized by dividing it by 200.}

\end{itemize}

\section{Numerical experiments}\label{exp}

Several numerical experiments using TGP are performed in this section by using the previously described multiobjective test problems. The general parameters of the TGP algorithm are given in Table \ref{tab1}. Note that the population size and the number of generations have been chosen as described in \cite{knowles1,zitzler1} in order to provide a fair comparison of the methods.

\begin{table}[htbp]
\caption{General parameters of the TGP algorithm for solving parity problems.}
\label{tab1}
\begin{center}
\begin{tabular}
{p{150pt}p{150pt}}
\hline
\textbf{Parameter}& 
\textbf{Value} \\
\hline
Population Size&
100\\
%\hline
Number of Generations&
250\\
%\hline
Insertion probability& 
0.05 \\
%\hline
Selection& 
Binary Tournament \\
%\hline
Function set & 
$+$, $-$, $*$, $sin$, $exp$ \\
%\hline
Terminal set& 
Problem variables \\
%\hline
Number of runs& 
30 \\
\hline
\end{tabular}
\end{center}
\end{table}

Because we are dealing with real-valued definition domains (e.g. [0,1]) we have to find a way of protecting against overflowing these. For instance if we employ a standard addition of two numbers greater than 0.5 we would get a result greater than 1 (domain upper bound). Thus, each operator has been redefined in order to output result in the standard interval [0,1]. The redefinitions are given in Table \ref{redefinitions}.

\begin{table}[htbp]
\caption{The redefinitions of the operators so that the output should always be between 0 and 1 if the inputs are between 0 and 1.}
\label{redefinitions}
\begin{center}
\begin{tabular}
{p{50pt}p{250pt}}
\hline
Operator& 
Redefinition \\
\hline
+& 
(x + y)/2 \\
%\hline
-& 
$\vert $x-y$\vert $ \\
%\hline
$*$& 
\textbf{None} (If you multiply two numbers between 0 and 1 you will always get a number between 0 and 1. \\
%\hline
sin& 
sin(x)/sin(1) \\
%\hline
exp& 
exp(x)/exp(1) \\
\hline
\end{tabular}
\end{center}
\end{table}

\begin{figure}[htbp]
\centerline{\includegraphics[width=\textwidth]{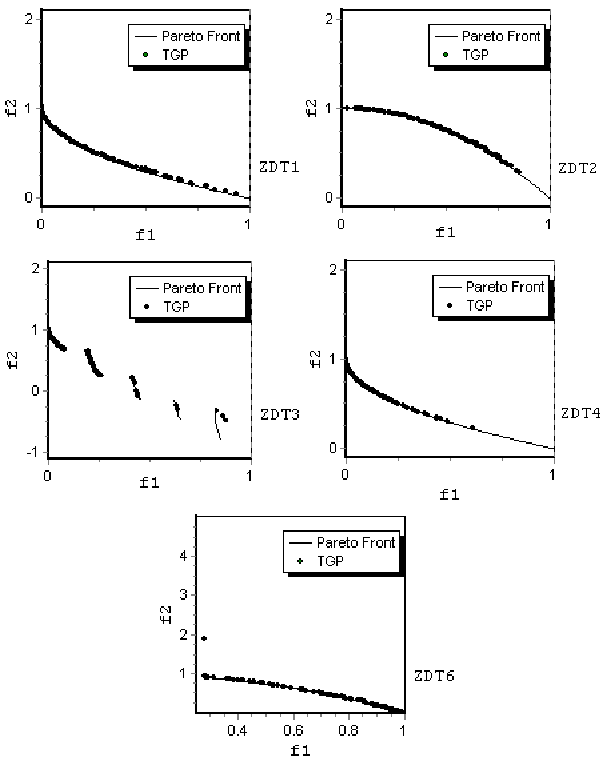}}
\caption{The results obtained by TGP without an archive for the test functions ZDT1-ZDT4, ZDT6. We have depicted the results obtained in one randomly chosen run (out of 30) in order to emphasize the poor distribution along the Pareto front.}
\label{figura3}
\end{figure}

Thirty independent runs have been performed. The results are depicted in Figure \ref{figura3}.

\begin{figure}[htbp]
\centerline{\includegraphics[width=\textwidth]{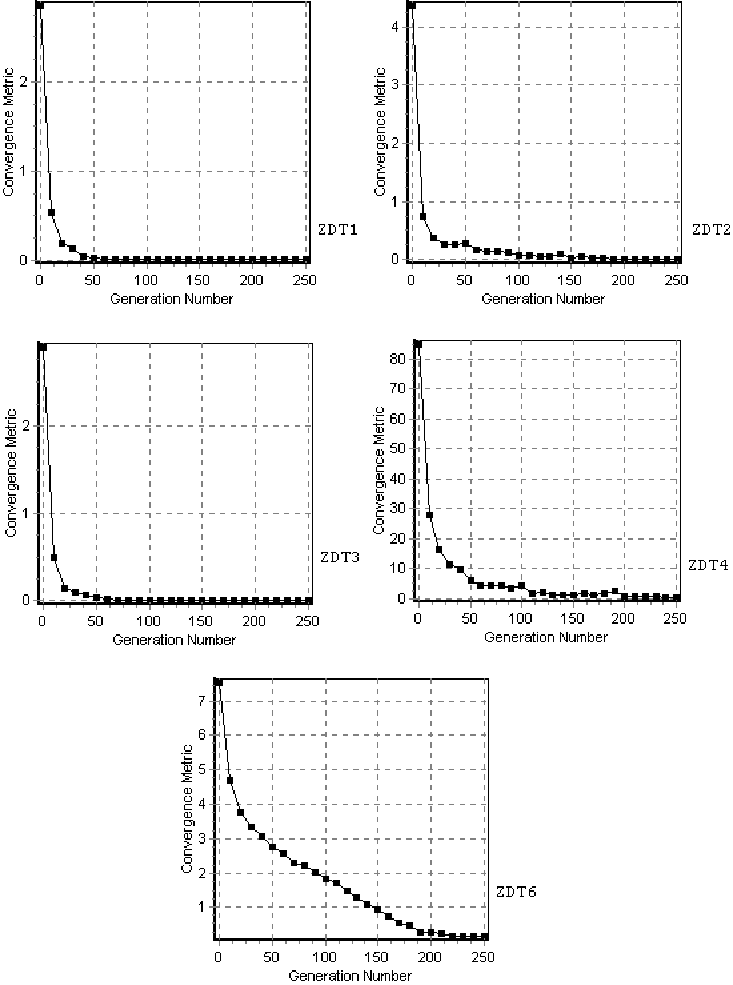}}
\caption{Convergence metric computed at every 10 generations. The results are averaged over 30 independent runs.}
\label{conv1}
\end{figure}

The convergence metric, computed at every 10 generations, is depicted in Figure \ref{conv1}. Note that only nondominated solutions have been taken into account for metrics computing.

\begin{figure}[htbp]
\centerline{\includegraphics[width=\textwidth]{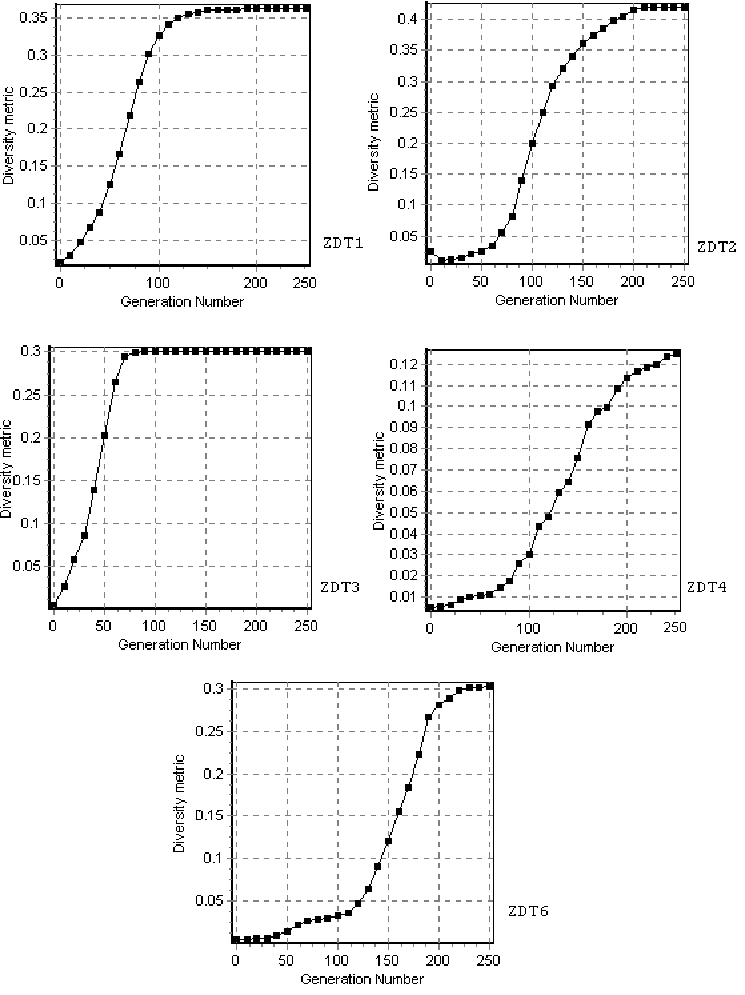}}
\caption{Diversity metric computed at every 10 generations. The results are averaged over 30 independent runs.}
\label{div1}
\end{figure}

The diversity metric, computed at every 10 generations, is depicted in Figure \ref{div1}.

Figures \ref{figura3}, \ref{conv1} and \ref{div1} show that TGP is able to provide a very good convergence towards the Pareto front for all the considered test problems (five). Unfortunately, the diversity of the solutions is sometimes very poor (see test functions ZDT2 and ZDT4). We have depicted the results obtained in a single run in order to provide a better view of the solution diversity. The same poor diversity has been observed in all other runs. However, the convergence of the solutions is very good, which makes us trust the power of the TGP method. This is why we have made a further step for improving this method, by adding a diversity preserving mechanism.

Numerical values of the convergence and diversity metrics for the last generation are also given in section \ref{comparison}.

\section{Improving diversity by using an archive}\label{tgp_archive}

The diversity of the obtained solutions is sometimes low (see Figure \ref{figura3}). The solutions tend to accumulate on the left side of the $Ox$ axis. This shortcoming could be due to the use of some of the operators given in Table \ref{redefinitions}. For instance, the result of the multiplication operator will always be lower than the value of its operands, thus generating a tendency toward the left side of the Ox axis. We have tried to remove this operator from the function set, but we have obtained even worse results. It seems that the set of operators used in conjunction with TGP could sometimes affect the quality of the obtained solutions.

In order to improve the diversity of solutions we have added to our algorithm, an archive which stores the nondominated solutions found so far. The archive is updated after each generation so that it contains the nondominated solutions from the new population and, of course, those from the old archive. If the number of the nondominated solutions exceeds the size of the archive the following steps are repeated until we obtain the required number of solutions in archive: the closest two solutions are computed and one of them is removed. This mechanism helps us maintain a reasonable diversity among the solutions in archive.

The selection for crossover is uniformly performed in the archive and in the current population. Individuals are randomly chosen with no emphasis on the nondominated ones. The nondominated individuals from the current population are not copied into the next population any longer.

\begin{figure}[htbp]
\centerline{\includegraphics[width=\textwidth]{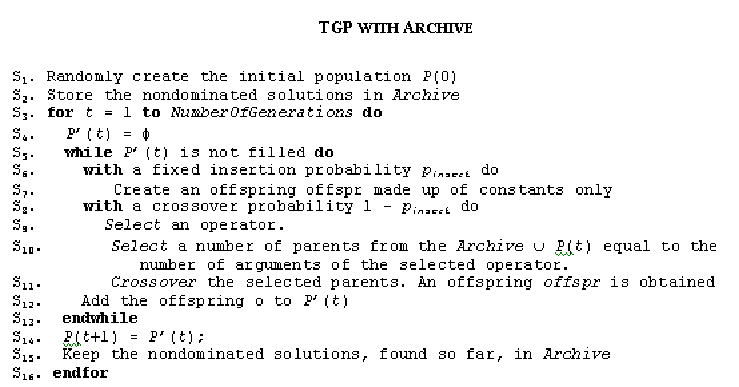}}
\caption{Traceless Genetic Programming with archive.}
\label{alg2}
\end{figure}

The algorithm is depicted in Figure \ref{alg2}.

\begin{figure}[htbp]
\centerline{\includegraphics[width=\textwidth]{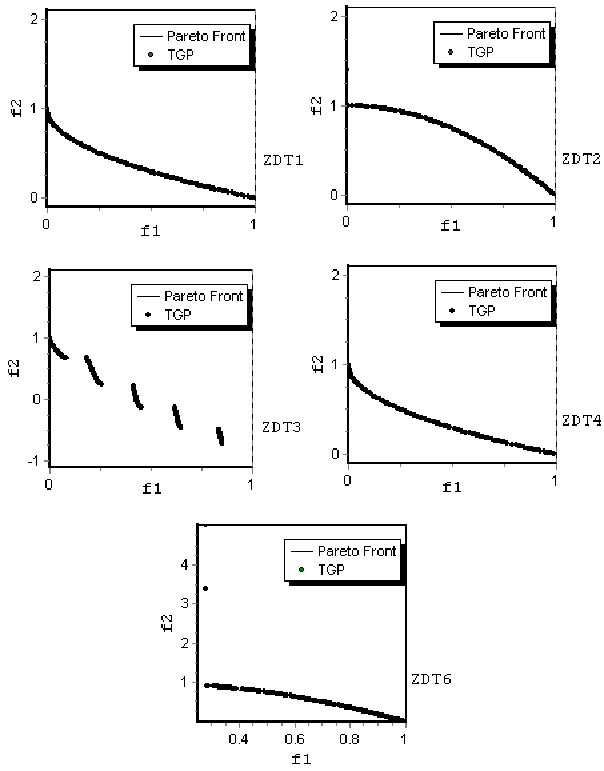}}
\caption{The results obtained by TGP with an archive for the test functions ZDT1-ZDT4, ZDT6. Only the results of a randomly chosen run have been depicted in order to provide a view of the solution diversity.}
\label{figura4}
\end{figure}

Thirty independent runs have been performed again for the test functions ZDT1-ZDT4, ZDT6. The results are depicted in Figure \ref{figura4}.

Figure \ref{figura4} shows that the algorithm achieved both a very good convergence and a very good diversity. The use of the archive has proven to be essential for preserving the diversity among the solutions.

\begin{figure}[htbp]
\centerline{\includegraphics[width=\textwidth]{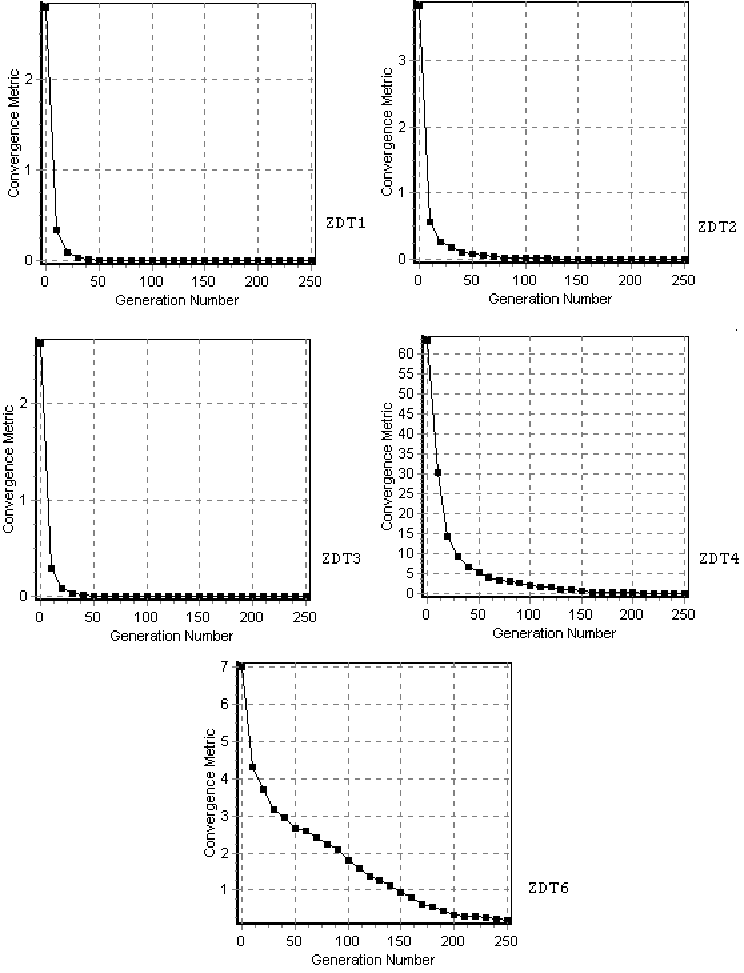}}
\caption{The convergence metric computed at every 10 generations for the TGP with archive. The results are averaged over 30 independent runs.}
\label{conv2}
\end{figure}

We have computed again the value of the convergence and diversity metrics in order to provide a better view of the newly introduced algorithm. The convergence metric, computed at every 10 generations, is depicted in Figure \ref{conv2}.

\begin{figure}[htbp]
\centerline{\includegraphics[width=\textwidth]{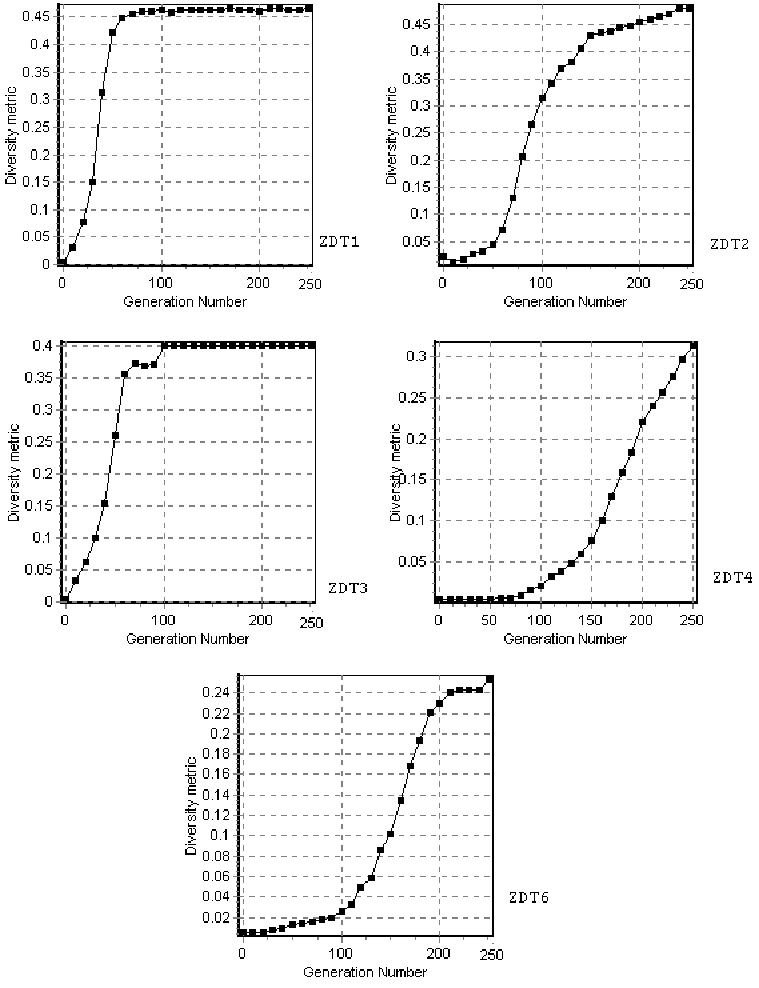}}
\caption{The diversity metric computed at every 10 generations for the TGP with archive. Results are averaged over 30 independent runs.}
\label{div2}
\end{figure}

The diversity metric, computed at every 10 generations, is depicted in Figure \ref{div2}.

Figure \ref{conv2} shows that the TGP-based algorithm quickly converges to the Pareto front. Less than 100 generations are required in order to obtain a very good convergence for the test functions ZDT1-ZDT3. A slower convergence is obtained for the test function ZDT6, thus proving the difficulty of this problem \cite{deb3}.

Numerical values of the convergence and diversity metrics for the last generation are also given in section \ref{comparison}.

\section{Running time}\label{running_time}

Table \ref{tab3} is meant to show the effectiveness and simplicity of the TGP algorithm by giving the time needed for solving these problems using a PIII Celeron computer at 850 MHz.

\begin{table}[htbp]
\caption{The average time for obtaining a solution for the multi-objective optimization problems by using TGP. The results are averaged over 30 independent runs.}
\label{tab3}
\begin{center}
\begin{tabular}
{p{50pt}p{140pt}p{120pt}}
\hline
Problem& 
Time (seconds) without archive&
Time (seconds) with archive\\
\hline
ZDT1& 
0.4&
19.1 \\
%\hline
ZDT2& 
0.4&
11.4 \\
%\hline
ZDT3& 
0.4&
7.5 \\
%\hline
ZDT4& 
0.3&
15.1 \\
%\hline
ZDT6& 
0.3&
3.2 \\
\hline
\end{tabular}
\end{center}
\end{table}

Table \ref{tab3} shows that TGP without archive is very fast. An average of 0.4 seconds is needed in order to obtain a solution for the considered test problems. However, the time needed to obtain a solution increases with at least one order of magnitude when an archive is used.

\section{Comparison with other methods}\label{comparison}

We compare the results obtained by TGP with the results obtained by two other well-known techniques: SPEA \cite{zitzler1} and PAES \cite{knowles1}. We have chosen these two methods because of their popularity in the field of MOEAs. Also, the results obtained by running these techniques (for the considered test problems) have been provided by the authors on the Internet \cite{paes_data,spea_data}. 

Results of the convergence and diversity metrics are given in Tables \ref{tab_conv} and \ref{tab_div}.

\begin{table}[htbp]
\caption{The convergence metric. The results are averaged over 30 independent runs. Lower values are better.}
\label{tab_conv}
\begin{center}
\begin{tabular}
{p{60pt}p{60pt}p{80pt}p{60pt}p{60pt}}
\hline
Problem& 
TGP&
TGP with archive&
SPEA&
PAES\\
\hline
ZDT1& 
0.010&
0.004&
0.039&
0.135\\
%\hline
ZDT2& 
0.006&
0.005&
0.069&
0.044\\
%\hline
ZDT3& 
0.005&
0.004&
0.018&
0.060\\
%\hline
ZDT4& 
0.331&
0.055&
4.278&
12.41\\
%\hline
ZDT6& 
0.161&
0.194&
0.484&
0.149\\
\hline
\end{tabular}
\end{center}
\end{table}

\begin{table}[htbp]
\caption{The diversity metric. The results are averaged over 30 independent runs. Higher values are better.}
\label{tab_div}
\begin{center}
\begin{tabular}
{p{60pt}p{60pt}p{80pt}p{60pt}p{60pt}}
\hline
Problem& 
TGP&
TGP with archive&
SPEA&
PAES\\
\hline
ZDT1& 
0.34&
0.465 &
0.299&
0.213\\
%\hline
ZDT2& 
0.419&
0.478 &
0.196&
0.213\\
%\hline
ZDT3& 
0.300&
0.399&
0.159&
0.151\\
%\hline
ZDT4& 
0.124&
0.312&
0.005&
0.005\\
%\hline
ZDT6& 
0.302&
0.253&
0.030&
0.153\\
\hline
\end{tabular}
\end{center}
\end{table}

Tables \ref{tab_conv} and \ref{tab_div} show that the TGP-based algorithm is able to solve the considered test problems very well, outperforming both SPEA and PAES. The only exception is at the ZDT6 test function where PAES converges better than TGP. The convergence of TGP is better with one order of magnitude than the convergence of SPEA and PAES on the test function ZDT4.

The diversity of the solutions produced by TGP is better than the diversity of SPEA and PAES for all test functions. In the case of test function ZDT4 the diversity of TGP is better with at least one order of magnitude than the diversity of PAES and SPEA.

The use of an archive has improved the diversity of TGP for the test functions ZDT1-ZDT4. TGP without archive has a better diversity than its counterpart with an archive for the test function ZDT6.

\section{Further work}

TGP with real-valued individuals has been applied (results not shown) for solving the multiobjective optimization problems \cite{deb2,khare} having a scalable number of objectives. The preliminary results indicated that real-valued encoding and the associated genetic operators described in this paper are not poweful enough for solving these problems better than other techniques. More sophisticated genetic operations should be used. 

We performed other experiments using binary encoding. In this case the results produced by our TGP algorithm have been significantly improved. They are competing the results produced by NSGA II \cite{deb4,khare}. More work should be done in this direction specially for tuning the parameters of the TGP. Also, binary representation \cite{holland1} requires more investigation within the TGP framework. 

A more detailed analysis of the operators used in conjunction with TGP will be performed in the near future. In this paper we have used some basic operators ($+, -, *, \slash, sin, exp$), but the choice of the function set may affect the quality of the obtained solutions. According to the No Free Lunch (NFL) theorems \cite{wolpert2} we cannot expect to find an optimal set of functions which will perform in the best way for all the optimization problems.

TGP will also be used for solving single objective \cite{yao1} and for solving equations.

\section{Conclusions}\label{conclusions}

Traceless Genetic Programming is a recent evolutionary technique which can be used for solving problems specific to GP and problems specific to GA. TGP uses a special individual representation, specific genetic operators and a specific evolutionary algorithm. 

In this paper, two TGP variants have been used for solving difficult multiobjective problems. Numerical experiments have shown that TGP is able to evolve a solution for the problems within a very short period of time. TGP-based algorithms have outperformed SPEA and PAES on the considered test problems.

\end{document}